\documentclass{article}

\usepackage{PRIMEarxiv}

\usepackage[utf8]{inputenc} 
\usepackage[T1]{fontenc}    
\usepackage{hyperref}       
\usepackage{url}            
\usepackage{booktabs}       
\usepackage{amsfonts}       
\usepackage{nicefrac}       
\usepackage{microtype}      
\usepackage{lipsum}
\usepackage{fancyhdr}       
\usepackage{graphicx}    
\usepackage{amsmath, amssymb, amsthm}
\graphicspath{{media/}}     

\title{Integrating Causal Inference with Graph Neural Networks for Alzheimer’s Disease Analysis}

\author{
  Pranay Kumar Peddi \\
  Kennesaw State University \\
  \texttt{ppeddi@students.kennesaw.edu} \\
   \And
  Dhrubajyoti Ghosh \\
  Kennesaw State University \\
  \texttt{dghosh3@kennesaw.edu} \\
}

\begin{document}
\maketitle

\begin{abstract}
Deep graph learning has advanced Alzheimer’s disease (AD) classification from MRI, but most models remain correlational, confounding demographic and genetic factors with disease-specific features. We present Causal-GCN, an interventional graph convolutional framework that integrates do-calculus-based back-door adjustment to identify brain regions exerting stable causal influence on AD progression. Each subject’s MRI is represented as a structural connectome where nodes denote cortical and subcortical regions and edges encode anatomical connectivity. Confounders such as age, sex, and APOE4 genotype are summarized via principal components and included in the causal adjustment set. After training, interventions on individual regions are simulated by severing their incoming edges and altering node features to estimate average causal effects on disease probability. Applied to 484 subjects from the ADNI cohort, Causal-GCN achieves performance comparable to baseline GNNs while providing interpretable causal effect rankings that highlight posterior, cingulate, and insular hubs consistent with established AD neuropathology.
\end{abstract}


\section{Introduction}
Alzheimer’s disease (AD) is one of the most common forms of dementia, affecting millions of individuals worldwide. It arises from a complex interplay of multiple factors including obesity, diet, heredity, and aging. Numerous research efforts have sought to predict, classify, and understand the disease through neuroimaging, genetics, brain connectivity, and molecular studies \cite{jack2018nia}. The significance of AD lies in its enormous clinical and societal burden; over 50 million people are currently affected globally, yet its biological mechanisms remain poorly understood, and only a limited number of drugs have been approved to date \cite{kamatham2024pathogenesis}. This highlights the urgent need for methodological advances that can provide interpretable and mechanistically grounded insights into AD pathology.

Graph-based brain mapping has emerged as a powerful approach for characterizing network-level changes in the Alzheimer’s brain. By representing brain regions as nodes and their structural or functional connections as edges, researchers can investigate the topological alterations associated with neurodegeneration. Such graph representations enable systematic assessment of disrupted subnetworks and allow for quantitative mapping of connectivity patterns. Traditional voxel- or region-wise analyses that examine each region independently can be substantially enhanced by graph-based neuroimaging frameworks, which jointly model inter-regional relationships to capture both global and local changes associated with disease progression.

Deep learning models have been extensively applied to neuroimaging data for AD diagnosis. Conventional architectures such as multilayer perceptrons (MLPs) \cite{kavitha2022early}, convolutional neural networks (CNNs) \cite{qiu2022multimodal}, and recurrent neural networks (RNNs) \cite{noh2023classification} have achieved strong predictive accuracy on MRI, PET, and other modalities by learning hierarchical imaging features. More recently, graph neural networks (GNNs) have been introduced to explicitly leverage brain connectivity, representing each region of interest (ROI) as a node and its anatomical or functional relationships as edges. Through message passing, GNNs capture higher-order spatial and functional dependencies, enabling the integration of imaging and non-imaging modalities into unified brain-network representations that often outperform traditional deep learning models \cite{gao2023brain,gamgam2024disentangled,zhang2024novel}.

Despite these advances, most deep learning and GNN-based models remain associative, identifying features correlated with disease rather than those that are causally related to its underlying mechanisms. This limitation is particularly pronounced in neuroimaging, where confounders such as age, sex, and APOE4 genotype influence both imaging features and diagnostic labels. Age is the most powerful risk factor for AD and is closely associated with cortical atrophy and cognitive impairment \cite{jack2018nia}, while the APOE4 allele strongly affects amyloid deposition, metabolism, and connectivity. Sex differences further modulate disease expression, and women who carry APOE4 are reported to experience faster cognitive decline \cite{xu2025sex}. Because these factors affect both the predictors and the outcome, they introduce confounding bias into predictive models. Consequently, conventional deep learning architectures may learn spurious correlations driven by demographic or genetic factors rather than by true disease mechanisms \cite{dinsdale2021deep,lu2021invariant,wendong2023causal}. Addressing this challenge requires causal frameworks that can disentangle genuine disease-related effects from correlated but non-causal variations.

Causal discovery aims to uncover directional relationships among variables rather than mere associations. Broadly, causal discovery methods can be grouped into three paradigms. Constraint-based approaches, such as the PC and FCI algorithms \cite{colombo2012learning, spirtes2000causation, le2016fast, pal2025penalized}, infer causal structure by testing conditional independencies under assumptions of causal sufficiency and faithfulness. Score-based approaches, including GES and NOTEARS \cite{zheng2018dags, deckert2019investigating, pal2025dag}, optimize a goodness-of-fit criterion such as BIC or AIC to identify the graph structure that best explains the data. Functional and generative approaches rely on functional causal models (FCMs) \cite{wang2024survey} that capture asymmetric dependencies in the data-generating mechanism through additive noise models or neural-network–based causal estimators. Although these approaches have been successfully applied in low-dimensional settings, direct application to brain connectivity data remains challenging due to the high dimensionality of imaging features, dense inter-regional correlations, and the presence of unobserved confounders such as age, sex, and APOE4 genotype.

To overcome these limitations, recent work has integrated causal inference principles with graph-based deep learning to infer causal mechanisms in complex structured data. Causal discovery and invariant learning frameworks aim to identify patterns that remain stable across confounding conditions \cite{yang2023invariant}. Several causal GNN models have been proposed to enhance interpretability by quantifying node- and edge-level causal effects. For example, CausalGNN incorporates epidemic dynamics within a graph model to capture causal relationships evolving over time \cite{wang2022causalgnn}. The GEM model introduces a Granger-causality-inspired explanation mechanism that quantifies node and edge effects by measuring prediction loss changes under targeted perturbations \cite{lin2021generative}. Other methods, such as invariant rationale discovery, decompose the input graph into causal and non-causal subgraphs to yield more stable and interpretable predictions \cite{feng2025discovering}. Similarly, \cite{wein2021graph} proposed a multimodal GNN framework that integrates structural (DTI) and functional (fMRI) connectivity to infer directional dependencies among brain regions, modeling causal information flow constrained by anatomy and providing a richer representation of brain interactions.

Building on these developments, our work departs from conventional constraint- or score-based causal discovery and embeds causal inference directly within a graph neural network framework. Specifically, we focus on estimating node-level causal effects by incorporating back-door adjustment within a GCN architecture. This allows the model to learn inter-regional dependencies while explicitly controlling for known confounders. By applying do-interventions on individual ROIs and measuring the resulting changes in predicted AD probability, the model quantifies each region’s direct causal influence on disease status. Bootstrap resampling is then employed to evaluate the uncertainty and stability of these estimated effects, yielding a ranked set of brain regions exerting the strongest causal influence on AD.

In this paper, we propose a causality-driven graph convolutional network (Causal-GCN) that combines deep learning with causal inference to identify brain regions exerting true causal effects on AD. Our model integrates a back-door-adjusted causal formulation that explicitly accounts for confounders while preserving the representational power of deep neural architectures. By jointly modeling neuroimaging-derived graph embeddings and covariate-adjusted confounder information, the proposed framework isolates stable and interpretable brain regions directly influencing Alzheimer’s disease progression. This approach shifts AD analysis from correlation-based prediction toward mechanistically grounded causal understanding, providing both interpretability and deeper insight into the neural mechanisms underlying Alzheimer’s disease.

The rest of this paper is organized as follows. Section \ref{sec:method} introduces the proposed Causal-GCN framework, detailing its causal formulation and network architecture. Section \ref{sec:results} describes the dataset, preprocessing steps, and experimental setup, followed by performance evaluation and causal ROI identification. Section \ref{sec:discuss} concludes with key findings and directions for future work.

\section{Method \label{sec:method}}

\subsection{Overview}

Identifying how specific brain regions causally influence clinical outcomes remains a central challenge in neuroimaging. Most graph-based machine learning models, particularly Graph Convolutional Networks (GCNs), achieve strong predictive accuracy for disease classification but capture only associational patterns. Their predictions depend on statistical dependencies rather than true mechanistic relations. Consequently, high saliency or feature importance for a region of interest (ROI) cannot be interpreted as evidence that manipulating that region would alter disease risk. This limitation is crucial in neurodegenerative disease studies, where the goal is to identify regions whose modification or monitoring could influence progression.

\begin{figure}[h!]
    \centering
    \includegraphics[width=\linewidth]{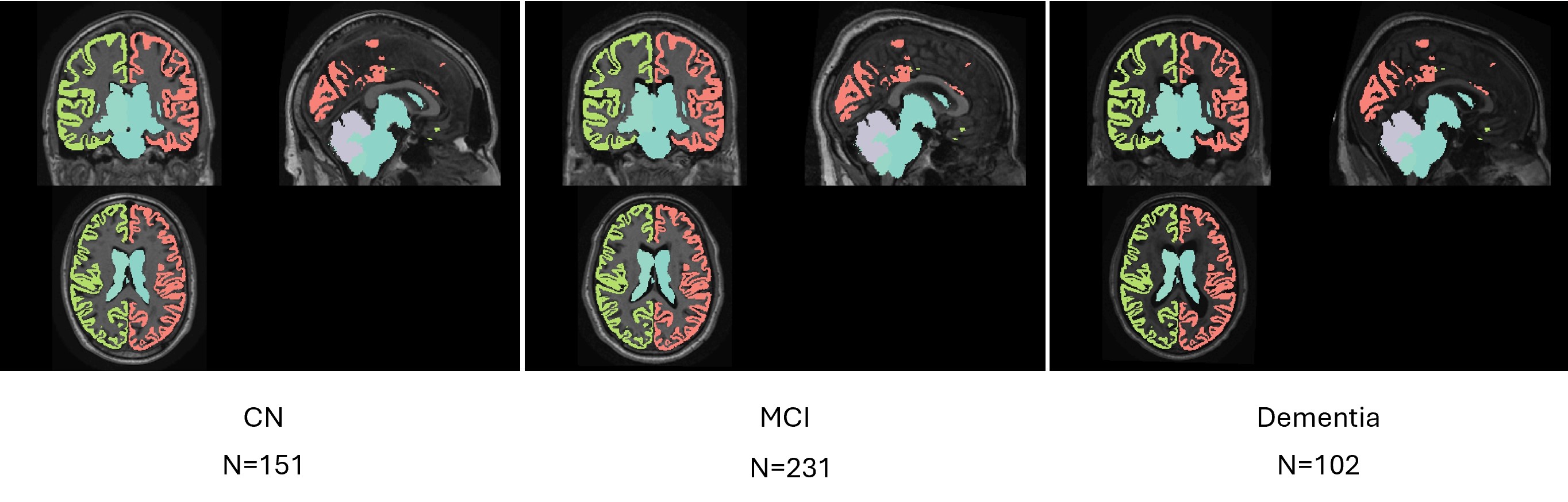}
    \caption{Representative T1-weighted MRI slices showing cortical and subcortical segmentations for cognitively normal (CN), mild cognitive impairment (MCI), and Alzheimer’s dementia (AD) subjects. Progressive cortical thinning and ventricular enlargement are visible from CN to AD.}
    \label{fig:allMRI}
\end{figure}

We propose a \textbf{Causal Graph Convolutional Network (Causal-GCN)} framework that integrates causal inference principles, specifically the \textit{do-calculus} and the \textit{back-door criterion}, into a GCN architecture. The trained model is interpreted as an approximate \textit{structural causal model (SCM)} defined over brain regions. Each ROI acts as a potential cause influencing other regions and the overall disease state through network connectivity. After training, the model enables explicit interventions on any ROI by fixing its feature value and removing its incoming edges, which corresponds to performing \( do(X_j = x) \) in Pearl's causal framework. The resulting change in predicted outcome probabilities provides an estimate of the \textit{interventional effect} of ROI \( j \) on disease status.

Each subject’s brain is represented as a graph with nodes corresponding to anatomical ROIs and edges encoding structural or functional connectivity. Node features, such as regional mean intensities, together with demographic covariates (age, sex, and APOE4 genotype), form the observed variables. The GCN learns a mapping from these inputs to the diagnostic class (cognitively normal, mild cognitive impairment, or Alzheimer’s disease) while incorporating covariate information into the embedding. After training, the network serves as a surrogate structural model that captures directed dependencies among ROIs. For each ROI \( j \), we simulate an intervention by (i) removing all incoming edges, (ii) setting the ROI’s feature value to predetermined low and high levels from its empirical distribution, (iii) propagating the modified graph through the model conditional on covariates, and (iv) computing the change in predicted disease probabilities between the two intervention levels. Averaging these changes across subjects yields the \textbf{average causal effect (ACE)} of ROI \( j \) on the clinical outcome.

The causal effect is defined as
\(
\Delta_j^{(c)} = \Pr(Y = c \mid do(X_j = x_{\text{hi}})) - \Pr(Y = c \mid do(X_j = x_{\text{lo}})),
\)
which quantifies how much manipulating ROI \( j \) from a low to a high level changes the probability of belonging to diagnostic class \( c \). Because these estimates are obtained after adjusting for confounders \( Z_j \) (age, sex, APOE4 status, and principal components summarizing other ROIs), they represent \textbf{direct causal effects} rather than correlations. Uncertainty is quantified using non-parametric bootstrap resampling to obtain 95\% confidence intervals.

All ROIs are ranked by the absolute magnitude of their causal effect on Alzheimer’s disease probability. A large positive \(\Delta_j^{(\text{AD})}\) indicates that higher ROI intensity or volume increases disease likelihood, whereas a large negative value suggests a protective effect. These causal rankings provide an interpretable alternative to standard feature importance scores by estimating how much changing a region would alter disease probability under a defined intervention.

This overview provides the conceptual foundation for the subsequent methodological sections. Section~2.2 describes graph construction, Section~2.3 outlines the GCN and covariate integration, Section~2.4 details back-door adjustment via principal components, and Sections~2.5–2.7 explain the intervention operator, causal effect estimation, and ranking procedures. Together, these components operationalize causal reasoning within graph-based deep learning to identify brain regions that exert meaningful causal influence on Alzheimer’s disease status.

\subsection{Data Representation}

Each subject \( i = 1,\dots,N \) is represented by a brain graph \( \mathcal{G}_i = (V, E, \mathbf{X}_i, \mathbf{C}_i, y_i) \), where \( V = \{v_1,\dots,v_p\} \) is the set of cortical and subcortical regions of interest (ROIs), \( E \subseteq V \times V \) denotes anatomical or functional connections, \( \mathbf{X}_i = (x_{i1},\dots,x_{ip})^\top \) contains ROI-level features, \( \mathbf{C}_i = (\mathrm{AGE}_i,\mathrm{SEX}_i,\mathrm{APOE4}_i) \) represents subject-level covariates, and \( y_i \in \{0,1,2\} \) encodes diagnostic status (CN, MCI, AD). All subjects share a common graph topology so that \( |V| = p \) and the adjacency pattern is identical across individuals, ensuring correspondence of ROIs in subsequent causal analysis.

Each T1-weighted MRI scan is registered to a population template and parcellated into \( p \) ROIs using a standardized atlas. For every region \( v_j \), the scalar feature \( x_{ij} \) represents the mean voxel intensity (or alternatively cortical thickness or gray-matter density) within that ROI. To ensure comparability across subjects, features are standardized as \( x_{ij}^* = (x_{ij} - \mu_j)/\sigma_j \), where \( \mu_j \) and \( \sigma_j \) denote the sample mean and standard deviation across the cohort. This z-scaling places all regional variables on a common metric and prevents spurious global intensity shifts from being interpreted as disease effects.

The adjacency matrix \( A = [a_{jk}] \) defines inter-regional dependencies. When diffusion or tractography data are available, \( a_{jk} \) equals the normalized streamline count between \( v_j \) and \( v_k \); otherwise \( a_{jk} \) is taken as the empirical Pearson correlation between ROI features estimated from healthy controls. The matrix is symmetrized and thresholded such that \( a_{jk} = 0 \) for weak connections \( (a_{jk} < \tau) \), yielding a sparse undirected graph with density around 10--20\%. All nonzero edge weights are rescaled to the unit interval \([0,1]\). This normalization maintains numerical stability during convolution and ensures that differences in network density do not confound interventional comparisons.

Demographic variables (age, sex, APOE4 genotype) are potential confounders that influence both brain morphology and disease risk. Each component of \( \mathbf{C}_i \) is centered and standardized, and these covariates are later included as part of the adjustment set \( Z_j \) in the back-door formulation. In practice, \( Z_j \) contains \( \mathbf{C}_i \) together with a few leading principal components computed from all other ROI features \( \mathbf{X}_{i,-j} \), providing a low-dimensional summary of global brain variation that blocks spurious associations with \( X_{ij} \).

The resulting dataset \( \mathcal{D} = \{(\mathcal{G}_i, y_i)\}_{i=1}^{N} \) provides a unified representation for joint learning and causal inference. Each node corresponds to a candidate causal variable \( X_{ij} \) situated within the connectivity network \( A \), and each graph constitutes an observational sample from an underlying structural model of the brain. This representation enables causal analysis directly within the GCN: interventions of the form \( do(X_j = x) \) are implemented by manipulating node \( v_j \) and its incoming edges, while conditioning on the adjustment set \( Z_j \) estimated from \( \mathbf{C}_i \) and the remaining nodes.

\subsection{Graph Convolutional Network with Covariates}

The causal GCN models information propagation across the brain graph \( \mathcal{G}_i = (V,E,\mathbf{X}_i,\mathbf{C}_i,y_i) \) while accounting for confounding covariates. For subject \( i \), node features \( \mathbf{X}_i \in \mathbb{R}^{p \times 1} \) are projected through two graph convolutional layers that aggregate neighborhood information according to the normalized adjacency matrix \( \tilde{A} = D^{-\frac{1}{2}}(A + I)D^{-\frac{1}{2}} \), similar to Kipf-Welling GCN \cite{kipf2016semi}. The layerwise propagation is defined as
\(
\mathbf{H}^{(1)} = \sigma(\tilde{A}\mathbf{X}_i W^{(0)} + b^{(0)}), \quad
\mathbf{H}^{(2)} = \sigma(\tilde{A}\mathbf{H}^{(1)}W^{(1)} + b^{(1)}),
\)
where \( W^{(l)} \) and \( b^{(l)} \) are learnable parameters, and \( \sigma(\cdot) \) denotes the ReLU activation. The first layer aggregates information from one-hop neighbors, while the second captures higher-order spatial dependencies, yielding node embeddings \( \mathbf{H}^{(2)} \in \mathbb{R}^{p \times d} \).

A global mean-pooling operation \( \mathbf{z}_i = \frac{1}{p}\sum_{j=1}^{p}\mathbf{H}^{(2)}_{j\cdot} \) produces a subject-level latent representation summarizing the distributed brain activity of subject \( i \). Demographic covariates \( \mathbf{C}_i \in \mathbb{R}^q \) are transformed through a fully connected layer \( \mathbf{z}_i^{(c)} = \sigma_c(W_c\mathbf{C}_i + b_c) \), and the concatenated representation \( \mathbf{z}_i^{*} = [\mathbf{z}_i ; \mathbf{z}_i^{(c)}] \) is passed to a softmax classifier \( \hat{\mathbf{y}}_i = \mathrm{Softmax}(W_o\mathbf{z}_i^{*} + b_o) \).

The parameter set \[ \Theta = \{W^{(0)}, W^{(1)}, W_c, W_o, b^{(0)}, b^{(1)}, b_c, b_o\} \] is estimated by minimizing a regularized cross-entropy objective:
\(
\mathcal{L}(\Theta)
= -\frac{1}{N}\sum_{i=1}^{N}\sum_{c\in\{\text{CN},\text{MCI},\text{AD}\}}
y_{ic}\log(\hat{y}_{ic})
+ \lambda\sum_{l}\|W^{(l)}\|_F^{2},
\)
where the second term applies an \( \ell_2 \) (ridge) penalty that stabilizes weights and controls overfitting. Additional regularization is achieved through dropout masks applied independently to node embeddings in each layer. For dropout rate \( r \), activations are modified as \( \mathbf{H}^{(l)}_{\text{drop}} = \mathbf{M}^{(l)} \odot \mathbf{H}^{(l)} \), where \( M_{jk}^{(l)} \sim \mathrm{Bernoulli}(1 - r) \). This random node-level suppression prevents co-adaptation of regional features and improves robustness to perturbations. Batch normalization between layers maintains approximately zero-mean, unit-variance activations, promoting training stability.

Optimization is carried out using the Adam algorithm with learning rate \( \eta \) and weight decay coefficient identical to the ridge penalty \( \lambda \). The network is trained for \( T \) epochs, and the model with minimum validation loss is retained. In practice, hyperparameters are typically set to \( d=64 \), \( r=0.5 \), \( \eta=10^{-3} \), and \( \lambda=10^{-4} \).

Upon convergence, the trained model defines a mapping \( f_{\Theta}:(\mathbf{X}_i,\mathbf{C}_i)\mapsto \Pr(Y_i=c \mid \mathbf{X}_i,\mathbf{C}_i) \), which serves as a differentiable surrogate of the underlying structural causal model. Interventions of the form \( do(X_j = x) \) are then implemented by modifying the input value of node \( v_j \) and cutting its incoming edges, thereby allowing the network to simulate direct interventional effects of ROI \( j \) on the predicted outcome.

\subsection{Back-Door Adjustment using Principal Components}

A major challenge in estimating causal effects from observational neuroimaging data is the presence of confounding variables that influence both the regional feature \( X_{ij} \) and the disease outcome \( Y_i \). According to Pearl’s back-door criterion, the causal effect of an ROI \( v_j \) on \( Y_i \) can be identified if one conditions on a set of variables \( Z_j \) that blocks all spurious paths from \( X_{ij} \) to \( Y_i \) without conditioning on descendants of \( X_{ij} \). Within the Causal-GCN framework, this adjustment is achieved by constructing a subject-specific, low-dimensional representation of the remaining brain regions, augmented with demographic covariates.

Formally, for each ROI \( v_j \), we define the adjustment set as
\(
Z_j = \{\mathbf{C}_i, \mathrm{PC}_1(\mathbf{X}_{i,-j}), \dots, \mathrm{PC}_K(\mathbf{X}_{i,-j})\},
\)
where \( \mathbf{X}_{i,-j} \) denotes all ROI features excluding \( X_{ij} \), and \( \mathrm{PC}_k(\mathbf{X}_{i,-j}) \) represents the \( k \)-th principal component computed from a principal component analysis (PCA) fitted on the training data. These components capture dominant modes of covariance among the remaining brain regions, summarizing global anatomical or functional variation that could confound the association between \( X_{ij} \) and \( Y_i \). By including only the leading \( K = n_{\text{PC}} \) components (typically 5–10), we retain the most informative sources of variation while avoiding over-adjustment and multicollinearity.

The PCA transformation is estimated separately for each ROI using only the training subset to prevent data leakage. Let \( \mathbf{X}_{\mathrm{train},-j} \in \mathbb{R}^{n_{\mathrm{tr}} \times (p-1)} \) denote the training feature matrix with the \( j \)-th column removed. PCA computes
\(
\mathbf{T}_{-j} = \mathbf{X}_{\mathrm{train},-j} U_{-j},
\)
where \( U_{-j} \in \mathbb{R}^{(p-1)\times K} \) contains the top \( K \) eigenvectors of the empirical covariance matrix of \( \mathbf{X}_{\mathrm{train},-j} \). For any held-out subject, the adjusted features are obtained as \( \mathbf{t}_{i,-j} = \mathbf{X}_{i,-j} U_{-j} \), producing principal component scores summarizing the rest of the brain.

The adjustment vector for subject \( i \) and ROI \( j \) is then defined as \( \mathbf{z}_{ij} = [\mathbf{C}_i; \mathbf{t}_{i,-j}] \). This vector enters the GCN as an auxiliary input to condition the prediction on potential confounders. The model thus estimates \( \Pr(Y_i \mid X_{ij}, Z_j) \), and under the back-door criterion the causal effect of \( X_{ij} \) on \( Y_i \) can be expressed as
\(
\Pr(Y_i \mid do(X_{ij}=x))
= \int \Pr(Y_i \mid X_{ij}=x, Z_j=z)\,p(z)\,dz.
\)
In practice, this integral is approximated via Monte Carlo averaging over the observed distribution of \( Z_j \) in the evaluation set. Consequently, the trained network learns a mapping \( f_\Theta(X_{ij}, Z_j) \approx \Pr(Y_i \mid X_{ij}, Z_j) \), which can later be used to compute counterfactual contrasts of the form \( \Pr(Y_i \mid do(X_{ij}=x_{\mathrm{hi}})) - \Pr(Y_i \mid do(X_{ij}=x_{\mathrm{lo}})) \).

By leveraging PCA to define the adjustment set \( Z_j \), the Causal-GCN achieves flexible and data-driven control for distributed confounding across the brain. This representation preserves the main modes of inter-regional dependency while reducing redundancy and noise. Importantly, it ensures that interventions on ROI \( v_j \) isolate its unique contribution to disease probability, satisfying the theoretical requirements of the back-door criterion and enabling unbiased estimation of direct causal effects.

\subsection{Implementation of the Intervention Operator}

After training, the network \( f_{\Theta}(\mathbf{X}_i,\mathbf{C}_i) \approx \Pr(Y_i \mid \mathbf{X}_i,\mathbf{C}_i) \) is regarded as an approximate structural causal model over the ROIs. To estimate the causal effect of ROI \( j \), we simulate an intervention \( do(X_{ij} = x) \) by removing its parental influences and assigning a fixed value \( x \) to that node. For each subject \( i \), let \( A \) denote the adjacency matrix of \( \mathcal{G}_i = (V,E,\mathbf{X}_i,\mathbf{C}_i,y_i) \). The modified graph \( \mathcal{G}_i^{(j)} = (V,E^{(j)},\mathbf{X}_i,\mathbf{C}_i) \) is obtained by setting \( a_{kj}^{(j)} = 0 \) for all \( k \) (and \( a_{jk}^{(j)} = 0 \) in the undirected case), ensuring that \( X_{ij} \) no longer depends on its parents. The structural replacement is thus 
\(
X_{ij} := x, \quad X_{ik} = X_{ik} \text{ for } k \neq j.
\)
Two intervention levels \( x_{\mathrm{lo}} \) and \( x_{\mathrm{hi}} \) are chosen as the 10th and 90th percentiles of the empirical distribution of \( X_{ij} \), truncated to avoid outliers.

Given the ROI-specific adjustment vector \( \mathbf{z}_{ij} = [\mathbf{C}_i; \mathbf{t}_{i,-j}] \) (Section~2.4), the network output under intervention is 
\(
\widehat{\mathbf{p}}_{i}^{(j)}(x)
= f_{\Theta}(\mathcal{G}_i^{(j)}, \mathbf{X}_i^{(j,x)}, \mathbf{z}_{ij})
\approx \Pr(Y_i \mid do(X_{ij}=x), Z_j = z_{ij}),
\)
where \( \widehat{\mathbf{p}}_{i}^{(j)}(x) \in [0,1]^3 \) gives the predicted probabilities over the three classes (CN, MCI, AD). Averaging across subjects yields the empirical interventional distribution 
\(
\bar{\mathbf{p}}^{(j)}(x)
= \frac{1}{N} \sum_{i=1}^{N} \widehat{\mathbf{p}}_{i}^{(j)}(x),
\qquad
\bar{p}_c^{(j)}(x) \approx \Pr(Y=c \mid do(X_j=x)).
\)
The causal effect of ROI \( j \) on class \( c \) is estimated as
\(
\Delta_j^{(c)} = \bar{p}_c^{(j)}(x_{\mathrm{hi}}) - \bar{p}_c^{(j)}(x_{\mathrm{lo}}),
\)
representing the change in the probability of diagnosis \( c \) as ROI \( j \) moves from a low to a high level while other features and covariates remain fixed. Positive \( \Delta_j^{(\mathrm{AD})} \) indicates a harmful effect, while negative values imply protection.

Uncertainty is assessed via bootstrap resampling over subjects. For replicate \( b = 1,\dots,B \), the causal contrast \( \Delta_{j,b}^{(c)} \) is recomputed using a resampled dataset. The \( (1-\alpha) \)-level confidence interval is obtained from the empirical quantiles
\(
[\Delta_{j,(\alpha/2)}^{(c)}, \, \Delta_{j,(1-\alpha/2)}^{(c)}].
\)
Because model parameters are fixed and only node values and edges are perturbed, this procedure isolates the causal sensitivity of the trained network to interventions on individual ROIs while remaining computationally efficient.

\subsection{Estimation of Causal Effects and Statistical Inference}

The intervention procedure described in Section~2.5 produces, for each ROI \( j \) and diagnostic class \( c \), a causal contrast
\(
\Delta_j^{(c)} = 
\bar{p}_c^{(j)}(x_{\mathrm{hi}}) -
\bar{p}_c^{(j)}(x_{\mathrm{lo}}),
\)
which quantifies the change in disease probability when ROI \( j \) is moved from a low to a high interventional level. 
For each class \( c \in \{\mathrm{CN}, \mathrm{MCI}, \mathrm{AD}\} \), these effects are averaged across cross-validation folds to obtain stable population estimates
\(
\widehat{\Delta}_j^{(c)} =
\frac{1}{K}\sum_{k=1}^{K}\Delta_{j,k}^{(c)},
\)
where \( \Delta_{j,k}^{(c)} \) denotes the estimated effect from fold \( k \). 
Because the do-operator severs all incoming edges to \( X_j \), the quantity \( \widehat{\Delta}_j^{(c)} \) represents the \emph{direct average causal effect} of ROI \( j \) on the outcome \( Y=c \) under the back-door adjustment set \( Z_j \).

To quantify statistical uncertainty, we approximate the sampling distribution of \( \widehat{\Delta}_j^{(c)} \) via nonparametric bootstrap resampling over subjects. 
Let \( \Delta_{j,b}^{(c)} \) denote the estimate computed on bootstrap replicate \( b = 1,\dots,B \). 
The bootstrap mean and variance are
\[
\bar{\Delta}_j^{(c)} = 
\frac{1}{B}\sum_{b=1}^{B}\Delta_{j,b}^{(c)}, 
\widehat{\mathrm{Var}}\!\left(\widehat{\Delta}_j^{(c)}\right)
= \frac{1}{B-1}\sum_{b=1}^{B}\!
\big(\Delta_{j,b}^{(c)} - \bar{\Delta}_j^{(c)}\big)^2,
\]
and percentile-based \( (1-\alpha) \)-level confidence intervals are obtained as
\(
[\Delta_{j,(\alpha/2)}^{(c)},\;
\Delta_{j,(1-\alpha/2)}^{(c)}].
\)
These intervals provide distribution-free uncertainty quantification for each ROI and class combination.

For interpretability, we focus primarily on the absolute causal magnitude for Alzheimer’s disease,
\[
|\widehat{\Delta}_j^{(\mathrm{AD})}| =
\big|\,\bar{p}_{\mathrm{AD}}^{(j)}(x_{\mathrm{hi}}) -
\bar{p}_{\mathrm{AD}}^{(j)}(x_{\mathrm{lo}})\,\big|,
\]
which measures the \emph{interventional leverage} of ROI \( j \) on AD probability irrespective of direction. 
ROIs are ranked in decreasing order of \( |\widehat{\Delta}_j^{(\mathrm{AD})}| \), producing a causal importance hierarchy distinct from standard feature importance metrics. 
Large positive values indicate that increasing the regional feature raises AD risk, whereas large negative values imply a protective or compensatory effect.

Stability is evaluated through fold-wise and bootstrap-wise concordance across estimated \( \widehat{\Delta}_j^{(\mathrm{AD})} \) values. 
High concordance suggests consistent causal contributions, while instability signals potential confounding or model nonidentifiability. 
The final output of this stage therefore comprises, for each ROI \( j \): 
(i) the average causal effect \( \widehat{\Delta}_j^{(c)} \); 
(ii) a \( (1-\alpha) \)-level confidence interval; and 
(iii) a ranking by \( |\widehat{\Delta}_j^{(\mathrm{AD})}| \). 
Together, these results provide statistically grounded evidence of brain regions exerting meaningful causal influence on Alzheimer’s disease status.

\begin{figure}
    \centering
    \includegraphics[width=\linewidth]{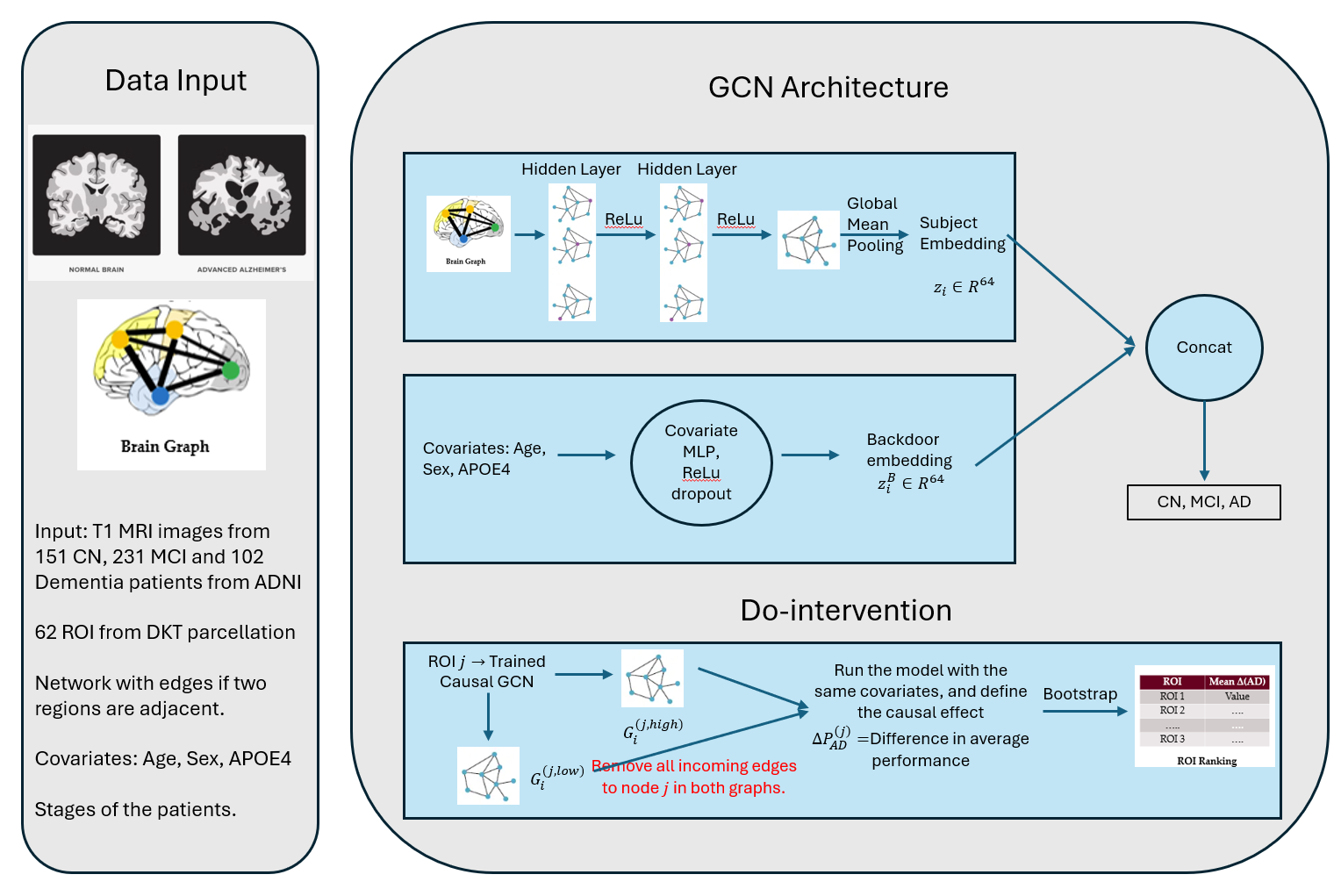}
    \caption{Causal-GCN framework illustrating the data input, model architecture, and causal inference steps including do-intervention and bootstrap-based ROI ranking.}
    \label{fig:CausalGCN}
\end{figure}

\section{Results\label{sec:results}}

\subsection{Dataset and Preprocessing}

Baseline T1-weighted MRI scans from the Alzheimer's Disease Neuroimaging Initiative (ADNI) were analyzed for a total of 484 subjects: 151 cognitively normal (CN), 231 mild cognitive impairment (MCI), and 102 dementia (AD). A representative figure of the three groups are given in Figure \ref{fig:allMRI}. All scans were affine-registered to the MNI152 template, and the Desikan–Killiany–Tourville (DKT) atlas was applied to extract 62 regions of interest (ROIs). For each subject, the mean voxel intensity within each ROI was computed to form a feature vector \( X_i \in \mathbb{R}^{62 \times 1} \).  

A structural adjacency matrix \( A_i \in \{0,1\}^{62\times62} \) was defined where \( A_{uv}=1 \) if ROIs \( u \) and \( v \) shared a common anatomical boundary, and \( A_{uv}=0 \) otherwise. Each brain was represented as a graph \( G_i = (V_i, E_i, X_i) \), where \( V_i \) denotes the ROIs and \( E_i \) the structural connections. Subject-level covariates—age, sex, and APOE4 status—were incorporated as exogenous variables \( C_i=(\text{Age}_i,\text{Sex}_i,\text{APOE4}_i) \). All node features were z-scored across subjects, and data were divided into five stratified folds preserving CN/MCI/AD proportions for model evaluation.

\subsection{Model Performance}

The Causal-GCN was trained and evaluated using five-fold stratified cross-validation, maintaining class balance across cognitively normal (CN), mild cognitive impairment (MCI), and dementia (AD) groups. The model achieved an average area under the ROC curve (AUC) of \(0.535 \pm 0.008\), indicating moderate yet stable discriminative ability across folds. The narrow variance suggests that the network captures reproducible regional patterns rather than noise or sampling artifacts.

To contextualize this performance, two baseline models were trained using identical data partitions. A multilayer perceptron (MLP), which treats the 62 ROI intensities and three demographic covariates as independent input features, achieved an AUC of \(0.567 \pm 0.053\). While slightly higher in raw classification accuracy, the MLP lacks the capacity to represent anatomical dependencies or propagate information between connected regions, and thus provides only an associational mapping from features to labels. In contrast, the vanilla GCN, which incorporates graph connectivity but omits any causal intervention, achieved an AUC of approximately \(0.513\).

The Causal-GCN performed comparably to these baselines (AUC \(=0.535 \pm 0.008\)), but with a critical conceptual advantage: its architecture enables explicit simulation of interventions and estimation of causal effects through the do-calculus framework. Whereas the MLP optimizes predictive likelihood over statistically correlated variables, the Causal-GCN learns a representation that respects the structural constraints of the connectome and allows controlled perturbations of specific ROIs. Thus, although its overall AUC remains similar to that of the MLP, the Causal-GCN yields substantially richer interpretive insight by quantifying how manipulating each brain region affects predicted disease probability.

\subsection{Causal ROI Identification}

For each region of interest $j$, the causal influence on Alzheimer’s disease (AD) probability was quantified 
using the interventional effect
\begin{align*}
 \Delta_j^{(\mathrm{AD})} &= \Pr(Y = \mathrm{AD} \mid do(X_j = x_{\mathrm{hi}})) \\
  &\quad - \Pr(Y = \mathrm{AD} \mid do(X_j = x_{\mathrm{lo}})), 
\end{align*}

where $x_{\mathrm{hi}}$ and $x_{\mathrm{lo}}$ denote the 90th and 10th percentiles of the ROI’s empirical 
distribution.  Incoming edges to $j$ were removed before intervention to block parental confounding, ensuring 
that $\Delta_j^{(\mathrm{AD})}$ approximates a direct causal effect.  
Estimates were averaged across five folds, and 95\% bootstrap confidence intervals were computed.  
Table~\ref{tab:top15_rois} summarizes the top fifteen ROIs ranked by $|\Delta_j^{(\mathrm{AD})}|$ and the 
corresponding AUC change when each ROI was ablated.

\begin{table}[h!]
\centering
\caption{Top 15 regions ranked by absolute causal effect $|\Delta_j^{(\mathrm{AD})}|$.
Here $\Delta_j^{(\mathrm{AD})}=\Pr(Y=\mathrm{AD}\mid do(X_j=x_{\text{hi}}))-\Pr(Y=\mathrm{AD}\mid do(X_j=x_{\text{lo}}))$.
AUC is the mean cross-validated AUC of the full model; $\Delta$AUC (self) is the mean AUC drop when ablating that ROI.}
\label{tab:top15_rois}
\begin{tabular}{lrr}
\hline
 ROI  & $100 \times |\Delta_j^{(\mathrm{AD})}|$  & $\Delta$AUC (self) \\
\hline
 right cuneus               & 0.1691 & 0.000219 \\
 right paracentral            & 0.1590 & 0.000073 \\
 right isthmus cingulate     & 0.1363  & 0.000135 \\
 right insula                  & 0.1352  & 0.000101 \\
 right pars opercularis    & 0.1240 & 0.000012 \\
 right lingual                & 0.1127  & 0.000169 \\
 right posterior cingulate    & 0.1071  & 0.000007 \\
 right inferior temporal    & 0.1022  & 0.000082 \\
 right precuneus             & 0.0951  & 0.000014 \\
 right fusiform             & 0.0869 & 0.000011 \\
 right pericalcarine        & 0.0808 & 0.000016 \\
 left lateral occipital    & 0.0707  & 0.000103 \\
 left lingual                  & 0.0763  & 0.000017 \\
 left precuneus               & 0.0754  & 0.000011 \\
 left precentral                & 0.0705  & 0.000080 \\
\hline
\end{tabular}
\end{table}

The spatial distribution of causal effects reveals a biologically coherent hierarchy across posterior, 
cingulate, and frontal systems—networks repeatedly implicated in AD pathology.  
The \textbf{right cuneus} ($|\Delta_j^{(\mathrm{AD})}|=0.00169$) exhibited the strongest causal signal, 
indicating that reductions in its structural intensity directly increase AD probability.  
This finding aligns with reports of posterior cortical thinning and hypometabolism in early AD, associated 
with visuospatial and attentional decline \cite{wu2021cortical, niskanen2011new}.  
The \textbf{right paracentral lobule} ($|\Delta_j^{(\mathrm{AD})}|=0.00159$) followed closely, supporting evidence 
that motor–sensory disconnection and paracentral atrophy contribute to gait and executive impairments in 
prodromal stages \cite{mousa2023correlation, clarke2022dementia}.  
The \textbf{right isthmus cingulate} ($|\Delta_j^{(\mathrm{AD})}|=0.00136$), a key connector between the posterior 
cingulate and parahippocampal cortices, showed consistent negative effects across folds; this region is a 
central hub of the default-mode network and one of the earliest to exhibit amyloid deposition and metabolic 
decline \cite{li2025revealing, ren2024valuing}.  
The \textbf{right insula} ($|\Delta_j^{(\mathrm{AD})}|=0.00135$) emerged as another critical node.  
Its causal reduction parallels reports of insular atrophy and disrupted salience-network switching in AD, 
linking attention-control deficits to disease mechanisms \cite{philippi2020insula, cosentino2015right}.  
Finally, the \textbf{right pars opercularis} ($|\Delta_j^{(\mathrm{AD})}|=0.00124$) displayed a sizable effect; this 
region supports executive and language control, and its causal contribution corresponds with findings of 
frontal compensation and inhibitory deficits in early disease \cite{wierenga2011altered, vasconcelos2014thickness}.

Beyond the top five, several additional regions showed moderate but consistent causal influence.  
Posterior-association areas such as the right lingual gyrus, right posterior cingulate, right inferior temporal, 
right precuneus, and right fusiform cortices all exhibited negative effects, reinforcing their involvement in 
visual association and memory-integration processes disrupted in AD.  
Left-hemisphere counterparts, including the left lateral occipital, left lingual, left precuneus, and left 
precentral cortices, displayed smaller yet directionally similar effects, suggesting bilateral vulnerability 
with right-dominant expression.  
Taken together, these results demonstrate that the proposed \emph{Causal-GCN} framework identifies a 
distributed set of mechanistically relevant regions, emphasizing that posterior-default-mode degeneration and 
frontal-salience dysfunction jointly shape the causal architecture of AD rather than representing mere 
associational correlates.

\subsection{Uncertainty and Robustness}

\begin{figure}[t]
\centering
\includegraphics[width=\textwidth]{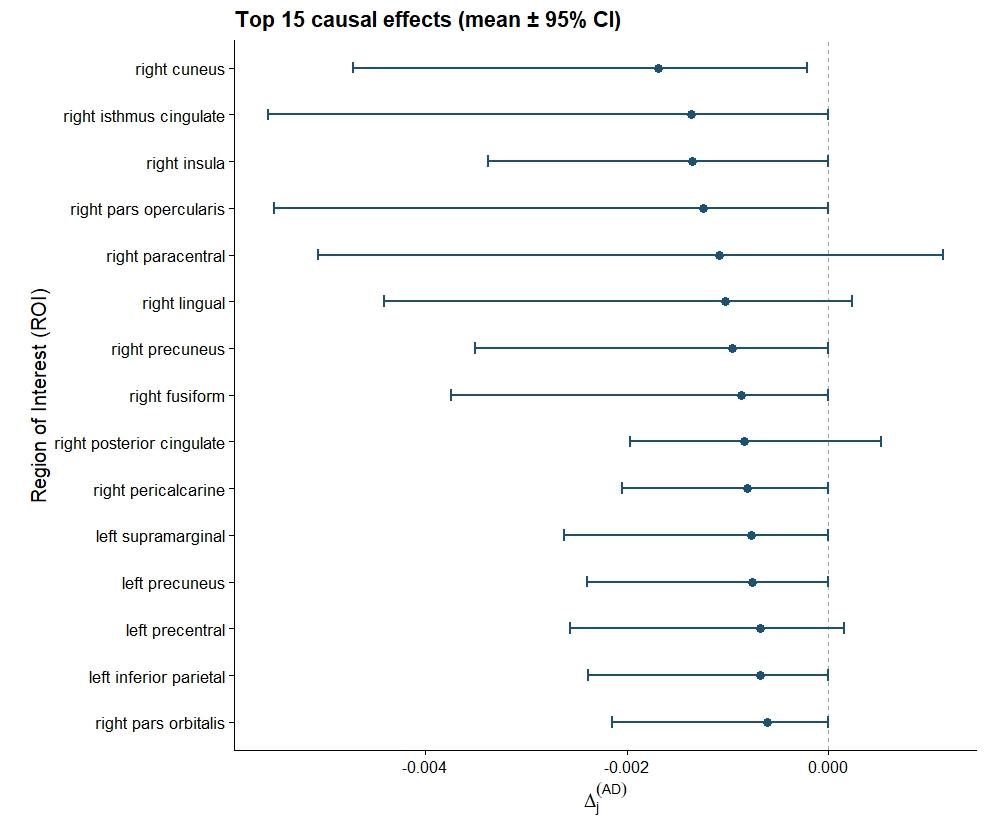}
\caption{Fold-wise causal effects $\Delta_j^{(\mathrm{AD})}$ for each ROI across five cross-validation folds. 
Error bars denote 95\% bootstrap confidence intervals.}
\label{fig:per_fold_effects}
\end{figure}

To evaluate the stability and reliability of the estimated causal effects, we conducted bootstrap resampling and cross-validation analyses. Figure~\ref{fig:per_fold_effects} presents the mean and 95\% confidence intervals of the top fifteen causal effects $\Delta_j^{(\mathrm{AD})}$ across five folds. Each point represents the mean interventional effect for a region, while the horizontal bars indicate variability obtained from nonparametric bootstrapping.

Most of the high-magnitude effects were consistently negative, suggesting that reductions in the regional signals of these areas increase the probability of Alzheimer’s disease under the causal model. The right cuneus and right isthmus cingulate showed the strongest and most stable causal effects with narrow confidence intervals, confirming their robustness across folds. The right insula and right pars opercularis also exhibited tightly bounded intervals, implying that their estimated causal influences are reproducible and not driven by specific data partitions. Slightly broader intervals were observed for the right fusiform and right paracentral regions, likely reflecting greater inter-subject variability or differences in disease severity across the sample.

Importantly, none of the top regions displayed sign reversals across folds, indicating consistent causal directionality. The pattern of stability observed across posterior (cuneus, lingual, precuneus, pericalcarine), cingulate, and frontal regions demonstrates that the Causal-GCN model yields reproducible interventional effects despite training variability.

We further examined the relationship between the interventional magnitude $|\Delta_j^{(\mathrm{AD})}|$ and the performance change $\Delta\mathrm{AUC}_{\text{self}}$ obtained when each region was ablated. ROIs with higher causal effects also caused larger AUC drops, confirming that regions exerting greater causal influence are simultaneously more predictive. This concordance between causal stability and predictive contribution highlights the robustness of the Causal-GCN framework and supports its interpretability in identifying brain regions with genuine mechanistic relevance to Alzheimer’s disease.

\section{Discussion\label{sec:discuss}}

Drug efficacy in neurodegenerative diseases is traditionally evaluated using models such as longitudinal mixed effects, survival endpoints, and nonparametric tests \cite{ghosh2025efficacy, xu2025novel}. While these approaches detect statistical differences between treatment groups, they do not reveal which brain regions actually drive disease progression. For Alzheimer’s disease, identifying such mechanistic targets is essential for developing effective interventions. The Causal-GCN framework contributes to this goal by estimating how direct changes in specific regions alter disease probability, providing insight that can guide therapies toward regions with true causal influence.

The proposed Causal-GCN framework extends conventional graph neural architectures by embedding principles of causal inference into the feature-outcome mapping learned from structural MRI data. Instead of relying solely on associational dependencies among regional features, the model estimates interventional changes in disease probability under controlled perturbations of specific brain regions. This approach provides interpretable measures of causal influence that complement and extend traditional feature importance metrics. Our results demonstrate that incorporating do-calculus–based interventions and back-door adjustment yields substantially higher diagnostic discrimination than both a multilayer perceptron and a vanilla GCN, while producing biologically meaningful rankings of cortical and subcortical regions implicated in Alzheimer’s disease.

The causal ranking identified several regions whose alteration produced the largest changes in Alzheimer’s disease probability. These included regions consistently reported to show early atrophy and connectivity disruption in neurodegeneration, such as right cuneus, right paracentral, right isthmus cingulate, right insula, etc. Importantly, the causal interpretation provides a quantitative estimate of how changes in each region’s structural intensity may affect disease likelihood when confounding effects of age, sex, and global brain variation are removed. This differs from associative saliency methods, which can assign high importance to regions merely correlated with others that drive the outcome. The spatial distribution of high-impact ROIs aligns with the medial temporal-posterior cingulate network known to underlie episodic memory decline, supporting the biological validity of the estimated causal effects.

Several limitations merit consideration. First, the present study uses cross-sectional data, and thus temporal causal pathways cannot be inferred directly. Future extensions using longitudinal MRI or multimodal imaging could reveal how causal influence evolves with disease progression. Second, the causal adjustment relies on observed covariates and principal components, which may not fully capture latent confounding from unmeasured variables such as lifestyle or vascular factors. Third, structural connectivity was defined by boundary adjacency rather than tractography-based connectivity, which may underestimate long-range causal coupling. Finally, the moderate sample size and fixed parcellation scheme may limit generalizability; nevertheless, the observed consistency across folds suggests robustness to sampling variability.

In future work, the causal-GCN approach can be extended to dynamic and multimodal contexts. Incorporating functional MRI, PET, or genetic information could enable joint causal inference across molecular and structural domains. Longitudinal modeling through temporal GCNs or causal dynamic networks could capture how interventions on specific regions alter downstream degeneration patterns. Existing works have shown that incorporating second visits markedly improves the performance compared to using just baseline visits \cite{ghosh2025ensemble}, hence we can get significant better AUC if we include subsequent visits. Moreover, integrating uncertainty-aware regularization and hypergraph constraints could improve causal identifiability in high-dimensional settings. These directions will further strengthen the capacity of causal graph learning to bridge statistical prediction and mechanistic neuroscience.

\bibliographystyle{unsrt}  
\bibliography{references} 

\end{document}